\documentclass[11pt,a4paper]{article}
\usepackage[hyperref]{acl2021}
\usepackage{times}
\usepackage{latexsym}
\usepackage{graphicx}
\usepackage{cellspace}
\usepackage{tabularx}

\usepackage{todonotes}

\usepackage{microtype}

\usepackage{colortbl}
\usepackage[T1]{fontenc}
\usepackage{color,soul}
\usepackage{booktabs}
\usepackage{multirow}
\usepackage{multicol}
\usepackage{diagbox}
\usepackage{booktabs}
\usepackage{amssymb}
\usepackage{pifont}
\usepackage{xspace}

\newcommand{\cmark}{\ding{51}\xspace}%
\newcommand{\xmark}{\ding{55}\xspace}%

\usepackage{linguex}
\usepackage{comment}

\usepackage{adjustbox}

\aclfinalcopy 
\def\aclpaperid{2765} 

\definecolor{ForestGreen}{RGB}{34,139,34}

\makeatletter
\newcommand{\printfnsymbol}[1]{%
  \textsuperscript{\@fnsymbol{#1}}%
}
\makeatother

\title{Figurative Language in Recognizing Textual Entailment}

\author{Tuhin Chakrabarty\thanks{~~Equal Contribution.}  \textsuperscript{1}, 
  Debanjan Ghosh\printfnsymbol{1} \textsuperscript{3}, 
  Adam Poliak\textsuperscript{2,4}
  \textbf{and} \textbf{Smaranda Muresan}\textsuperscript{1,4}\\ 
  \textsuperscript{1}Department of Computer Science, Columbia University \\
  \textsuperscript{2}Department of Computer Science, Barnard College \\
  \textsuperscript{3}Educational Testing Service,
  \textsuperscript{4}Data Science Institute, Columbia University\\
  {\tt \{tuhin.chakr, smara\}@cs.columbia.edu}, \\{\tt dghosh@ets.org}, {\tt apoliak@barnard.edu}
  }

\date{}

\begin{document}
\maketitle
\begin{abstract}
We introduce a collection of recognizing textual entailment (RTE) datasets focused on figurative language. We leverage five existing datasets annotated for a variety of figurative language -- simile, metaphor, and irony -- and frame them into over 12,500 RTE examples.We evaluate how well state-of-the-art models trained on popular RTE datasets capture different aspects of figurative language. Our results and analyses indicate that these models might not sufficiently capture figurative language, struggling to perform pragmatic inference and reasoning about world knowledge. Ultimately, our datasets provide a challenging testbed for evaluating RTE models.

\end{abstract}

\section{Introduction}
Figurative language is ubiquitous in many forms of discourse from novels, poems, and films, to scientific literature and social media conversations \cite{ghosh2018verbalirony}.
It is often used to convey intimacy~\cite{gerrig1988beyond}, humour~\cite{roberts1994people}, intense emotions~\cite{fussell1998figurative}, or veiled politeness~\cite{jorgensen1996functions}. Despite its ubiquity, figurative language remains ``a  bottleneck in automatic text understanding''~\cite{shutova2011computational}. 

\begin{table}[t!]
 \centering
 \small
 \renewcommand{\arraystretch}{1.25}
 \begin{tabular}{p{1cm}p{5.45cm}|p{.01cm}}
     \toprule 
& $\blacktriangleright$ 
I start to prowl across the room like a \textbf{tightrope walker on dental floss}. \\ 
 &  \hspace{1em}I start to prowl across the room \textit{recklessly}. & 
\multirow{-2}{*}{\xmark} \\ 
 & $\blacktriangleright$ 
 They had shut him in a basement that looked like a \textbf{freight elevator}. &  \\ 
\multirow{-4}{*}{Simile} & \hspace{1em}They had shut him in a basement that looked \textit{dangerously claustrophobic}. & \multirow{-2}{*}{\cmark}  \\
\midrule 
& $\blacktriangleright$ He \textbf{weathered} the costs for the accident. & \\
 & \hspace{1em} He \textit{avoided} the costs for the accident. & \multirow{-2}{*}{\xmark} \\ 
\multirow{-2}{*}{Metaphor} & $\blacktriangleright$ The bus \textbf{bolted} down the road.  & \\  
& \hspace{1em}The bus \textit{paced} down the road. & \multirow{-2}{*}{\cmark} \\ \midrule 
 
 & $\blacktriangleright$ 
  Made \$174 this month, gonna buy a yacht!  &  \\ 
 & \hspace{1em}I don't make much money.  &
 \multirow{-2}{*}{\xmark}  \\
 Irony & $\blacktriangleright$ 
 Fans seem restless, gee, don't understand them.  &  \\ 
 & \hspace{1em} Fans seem restless - don't know the reason behind it.  &
 \multirow{-2}{*}{\cmark}  \\
\bottomrule 
 \end{tabular}
 \caption{Example RTE pairs focused on similes, metaphors, and irony that RoBERTa \emph{incorrectly} labels. $\blacktriangleright$ indicates a context and the following sentence is its corresponding hypothesis.
\cmark and \xmark respectively indicate that the context entails, or does not entail the hypothesis. \textbf{Bold} text represent simile and metaphors and \emph{Italic} represent their entail/not entail interpretations (top two rows).}


 \label{tab:nli_examples}

 \end{table}

Recognizing Textual Entailment (RTE), the task of identifying whether one sentence (context) likely entails another (hypothesis), is often used as a proxy to evaluate how well Natural Language Processing (NLP) systems understand natural language~\cite{cooper1996using,dagan2006pascal,bowman2015large}. Figurative language is defined as any figure of speech which depends on a non-literal meaning of some or all of the words used. Thus, understanding figurative language can be framed as an RTE task (figurative language expression vs. intended meaning), where the figurative language expression is the \emph{context} and the intended meaning is the \emph{hypothesis} in an RTE framework (See examples in ~\autoref{tab:nli_examples}).

We investigate how suitable are state-of-the-art RTE models trained on current RTE datasets to capture figurative language.   
We focus on three specific types of figurative language: 
similes, metaphors, and irony. Similes evoke comparisons between two seemingly different objects, metaphors expand the imagination beyond the literal narrative, and irony conveys the opposite
of what is said. 

We leverage five existing datasets annotated for these types of figurative language
to create over 12,500 RTE examples that require understanding or identifying these phenomena. 
We evaluate how well standard neural RTE models capture these aspects of figurative language. Our results demonstrate that, although, systems trained on a popular RTE dataset 
may capture some aspects of various types of figurative language, they fail on cases where the interpretation relies on pragmatic inference and reasoning about world knowledge. We release the code and the data.
\footnote{\url{https://github.com/tuhinjubcse/Figurative-NLI}}

\section{Related Work} \label{section:relatedwork}
We follow recent work that test for an expanded range of inference patterns in RTE systems~\cite{bernardy2019kind} by
evaluating how well RTE models capture specific linguistic phenomena, such as  pragmatic inferences~\cite{jeretic-etal-2020-natural}, veridicality~\cite{ross-pavlick-2019-well}, and others~\cite{P16-1204,white2017inference,2018arXiv180204302D,Naik2018,glockner-shwartz-goldberg:2018:Short,kim-etal-2019-probing,kober2019temporal,Richardson2020ProbingNL,yanaka2020neural,Vashishtha-EMNLP-20,poliak2020survey}.

We are not the first to explore figurative language in RTE. \newcite{agerri-2008-metaphor} analyze examples in the Pascal RTE-1~\cite{dagan2006pascal} and RTE-2~\cite{BarHaim2006TheSP} datasets that require understanding metaphors and \newcite{agerri-etal-2008-textual} present an approach for RTE systems 
to process metaphors. 
\newcite{poliak-etal-2018-collecting-diverse}'s
diverse collection of RTE datasets includes examples based on figurative language, but focuses only on identifying puns.

\section{Dataset Creation} \label{section:dataset}
We create RTE test sets that focus on similes, metaphors, and irony.   
We provide further background for these types of figurative language 
and describe the methods used for creating these test sets. \autoref{table:stat} reports the final test sets' statistics.

\begin{table}[]
\centering
\small
\begin{tabular}{|c|c|c|c|c|}
\hline

\multicolumn{2}{|c|}{Data}     & Total & E & NE \\ \hline
\multicolumn{2}{|c|}{Simile}   &  600   &  300 & 300   \\ \hline
\multicolumn{2}{|c|}{Metaphor} &  613    & 307  & 306   \\ \hline
\multirow{2}{*}{Irony Meaning}  & $SIGN_{2000}$ & 2,000      & 133  & 1867   \\ \cline{2-5}
& S$_{im}$-H$_{int}$ & 4,762   & - & 4,762 \\ \hline
\multicolumn{2}{|c|}{Irony  Intention}  &   4,601    &  2,212 & 2,389   \\ \hline
\end{tabular}
\caption{\label{table:stat}
Dataset statistics and class distribution, \emph{Entailment} (E) and \emph{Not-Entailment} (NE) for each type of figurative language.
}
\end{table}

\fi

\subsection{
Simile}
Comparisons are inherent linguistic devices that express the likeness of two entities, concepts, or ideas. When used figuratively, comparisons are called similes. 
Similes are used to spark the reader's imagination by making descriptions more emphatic or vivid~\cite{definition}. Similes use a common PROPERTY to compare two concepts often referred to as the TOPIC (the logical subject) and the VEHICLE (the logical object of comparison). For example, in the simile ``Love is like an unicorn'', love (TOPIC) is compared to a unicorn (VEHICLE), portraying the implicit property ``rare''. Recently \citet{chakrabarty-etal-2020-generating} released a test set of 150 literal sentences from subreddits r/WritingPrompts and r/Funny, each aligned with two human-written paraphrases with similes that retain the original meaning. 

To create our RTE test set that focuses on similes, we treat these simile-literal aligned sentences as entailed context-hypothesis pairs. Given a literal input, ``{They  had  shut  him  in  a  basement  that looked \textbf{dangerously claustrophobic}}", an expert annotator re-framed it as ``{They  had  shut  him  in  a  basement  that looked \emph{like a freight elevator}}".\footnote{Note, such re-framing task (content generation task) does not involve assigning a label to a text fragment, thus, computing inter-annotator agreement is not applicable here.} We create Not-Entailed examples by flipping the literal verb/property with their respective antonyms and use the original (Literal, Simile) pairs as Entailed.
For instance, in the case of an existing context-hypothesis pair expressing \emph{Entailment} - ``An ordinary citizen coming to power in this way is like \textbf{a green moon.}'' $\rightarrow$ ``An ordinary citizen coming to power in this way is \emph{unprecedented}" -  we alter ``unprecedented" to ``common" to make it a pair of \emph{Not-Entailment} (NE) instance. 

\begin{table*}[]
     \renewcommand{\arraystretch}{1.25}
    \centering
    \begin{adjustbox}{width=1\textwidth}
    \begin{tabular}{c|c|l}
\toprule
Genre                      & PairID                   & \multicolumn{1}{l}{Example}                                                                                                                               \\ \midrule
\multirow{3}{*}{Slate}     & \multirow{3}{*}{143311e} & $\blacktriangleright$ Praise from a stranger is \textbf{like a glass of water} served at a restaurant in: You drink it warily, if at all, \\
                           &                          & fearing it may be tainted                                                                                                                                  \\
                           &                          & \hspace{1em}Praise from someone you do not know can be taken lightly                                                                      \\ \cline{1-3}
\multirow{2}{*}{Fiction}   & \multirow{2}{*}{60838c}  & $\blacktriangleright$ The stars are no more like the sun than the glow of my cigarette is \textbf{like a forest fire}.                    \\
                           &                          & \hspace{1em}The sun is comparable to the stars because they are the same.                                                                 \\ \cline{1-3}
\multirow{3}{*}{Telephone} & \multirow{3}{*}{99298c}  & $\blacktriangleright$ But uh still I I question the ability of some of the teachers to uh really do a \textbf{bang-up job} and            \\
                           &                          & yet others i know are just wonderful                                                                                                                       \\
                           &                          & \hspace{1em}All teachers sucks                                                                                                            \\ \bottomrule
\end{tabular}
    \end{adjustbox}
    \caption{\label{tab:mnli-examples}Examples from MNLI that include figurative language. $\blacktriangleright$ indicates a context and the following line is its corresponding hypothesis.}
\end{table*}

\subsection{
Metaphor}
Metaphors express deep feelings and complex attitudes~\cite{veale2016metaphor}. Understanding metaphors requires comprehending abstract concepts and making connections between seemingly unrelated ideas to appropriately deviate from literal meaning \cite{gutierrez2016literal,mohammad-etal-2016-metaphor,kintsch2002metaphor,glucksberg1998understanding}.When generating metaphoric paraphrases, \citet{chakrabarty2021mermaid} create a diverse test set of 150 literal sentences curated from different domains and genres and asked two expert annotators to create metaphorical sentences, resulting in  a total of 300 metaphorical examples. The expert annotators re-framed the literal sentences independently by replacing the literal verb with a metaphorical verb. For instance, an expert reframed the literal sentence ``{The tax cut will help the economy}" to ``{The tax cut will \textbf{fertilize} the economy}".

Since the most frequent type of metaphor is expressed by verbs~\cite{martin2006corpus,steen2010method} these literal and metaphorical paraphrases differ only by the verb they use. In an RTE framework, we treat these metaphorical-literal pairs as entailed context-hypothesis examples.
To create Not-Entailed examples, we generate hypotheses by manually swapping the literal verb in the entailed hypothesis with its antonym.
\begin{table*}[t!]
\small
\centering
\begin{tabular}{|p{2.5cm}|p{2cm}|p{2cm}|p{2cm}|p{2cm}|p{2cm}|}
\hline
\multirow{2}{*}{\diagbox[width=7.5em]{Model}{Testset}} & \multirow{2}{*}{Simile} & \multirow{2}{*}{Metaphor} & \multicolumn{2}{c|}{IMeaning}  & \multirow{2}{*}{IIntent} \\ \cline{4-5}
& &  & $sm-im$ & $SIGN_{2000}$ & 
\\ \hline
NBoW      &  51.17   &   54.81  &  86.37 &  71.50    &  \textbf{61.72} \\ \hline
InferSent & 55.01   &   65.75  &  71.62 &      68.84 &  11.72\\ \hline
RoBERTa-large   & \textbf{85.47}   &   \textbf{88.09}  &  \textbf{94.76} &  \textbf{93.42}   &  52.81 \\ \hline
\end{tabular}
\caption{Accuracy of different models on our datasets focusing on similes, metaphors, and irony.}
\label{table:resultsall}
\end{table*}
Note that for both simile and metaphor, automatic substitution using available lexicons is problematic as it often leads to ungrammatical sentences. Manually replacing the words with its antonym guarantees a high quality test set. We use antonyms to create Not-Entailed examples for Simile and Metaphors which contain both Neutral and Contradiction classes. Such lexical replacement using antonyms would clearly lead to higher quality contradiction example creation. On the contrary, creating neutral examples by lexical perturbation is challenging and if not done properly, it can lead to grammatical errors or incoherent sentences. 

\subsection{
Irony}
When using irony, speakers usually mean the opposite of what they say~\cite{sperber1981irony,dews2007not}. We develop different test sets focusing on whether the RTE models should \textit{understand the conveyed meaning} of ironic examples or should \textit{identify the speaker's ironic intent} (i.e., identify if an utterance is ironic or not) given the hypothesis that the speaker was ironic. 

\paragraph{Understanding Ironic Meaning (IMeaning)}

\newcite{peled2017sarcasm} used skilled annotators to create a parallel dataset between tweets with verbal irony and their non-ironic rephrasings (15K pairs). Annotators also had the option to copy the original tweet or just to paraphrase it, in case the ironic intent is not easy to identify. Likewise, \newcite{ghosh-etal-2020-interpreting} released a parallel dataset of speakers' ironic messages (S$_{im}$) and hearers' interpretations (H$_{int}$) of the speaker's intended meaning. This dataset (S$_{im}$-H$_{int}$) contains 4,761 ironic-literal pairs.
We use both datasets in our experiments and henceforth denote them as $SIGN$ and S$_{im}$-H$_{int}$,  respectively. For both datasets, the original \emph{ironic} messages are treated as the contexts and the \emph{intended} meanings are the hypotheses. However, all RTE contexts do not contradict their corresponding hypotheses. For instance, in case of \newcite{peled2017sarcasm}, the authors allowed annotators to not rephrase the ironic sentences with their opposite \emph{intended meanings} (in case the sarcastic or ironic intent was not clear). Thus, for evaluation purposes (see Table \ref{table:resultsall}), we annotated a subset of 2,000 random pairs from $SIGN$ and evaluated the RTE models on that subset (denoted as $SIGN_{2000}$ henceforth). Around 93\% of the RTE pairs in $SIGN_{2000}$ are Not-Entailed examples and 100\% of RTE pairs in S$_{im}$-H$_{int}$ are Not-Entailed examples.

\paragraph{Recognizing Ironic Intent (IIntent)}
We leverage additional ironic examples from \newcite{van2018semeval}. Following \newcite{poliak-etal-2018-collecting-diverse}'s method for recasting annotations for puns and sentiment, we use \emph{templates} to generate contexts (a)  and hypotheses (b). We use all the ironic tweets ($training$ and $test$) released by \newcite{van2018semeval} to generate 4,598  
 RTE pairs. Akin to \newcite{poliak-etal-2018-collecting-diverse}, we replace \emph{Name} with names sampled from a distribution of names based on the US census data.\footnote{http://www.ssa.gov/oact/babynames/names.zip}. The templates are 
a) \textit{Name} tweeted that \textit{tweet}, b) \textit{Name} was ironic.

\section{Experimental Setup} 

MNLI~\cite{williams2017broad} is one of the widely used large-scale corpora that contains instances of figurative language (Table \ref{tab:mnli-examples}). Following recent work, we evaluate RTE models trained on MNLI~\cite{williams2017broad} using three standard neural models: bag of words (NBoW) model, InferSent \cite{conneau-EtAl:2017:EMNLP2017}, and RoBERTa-large \cite{liu2019roberta}. In NBoW, word embeddings for contexts and hypotheses are
averaged separately, and their concatenation is passed to a logistic regression softmax classifier. InferSent encodes the context and hypotheses independently using a BiLSTM, then their sentence representations are fed to a MLP.\footnote{Both NBoW and InferSent use $300$D Glove embeddings~\cite{pennington2014glove}.}
For RoBERTa, we use the model fine-tuned on MNLI from the Transformer's library~\cite{wolf-etal-2020-transformers}.
We expect models trained on MNLI to capture some forms of figurative language that
often appear in works of fictions, conversations,  speeches, and magazines like Slate. Table \ref{tab:mnli-examples} illustrates a few examples from MNLI that include figurative language

\section{Results and Discussions} \label{subsection:results}

\begin{table*}[t]
\renewcommand{\arraystretch}{1.25}
\small
\centering
\begin{tabular}{|l|l|l|l|l|}
\hline
                          &                    &                                                                                                               & Gold                & Pred                \\ \hline
\multirow{6}{*}{Simile}   & \multirow{2}{*}{$\blacktriangleright$} & Your guardian angel is just a little too much like a \textbf{nerd at a comic convention}.    & \multirow{2}{*}{\cmark}  & \multirow{2}{*}{\xmark} \\ 
                          &                    & Your guardian angel is just a little too \textit{enthusiastic}                               &                     &                     \\ \cline{2-5} 
                          & \multirow{2}{*}{$\blacktriangleright$} & Growing up, people always thought you were like a \textbf{social pariah}.                    & \multirow{2}{*}{\xmark} & \multirow{2}{*}{\cmark}  \\ 
                          &                    & Growing up, people always thought you were \textit{ordinary}                                 &                     &                     \\ \cline{2-5} 
                          & \multirow{2}{*}{$\blacktriangleright$} & They all agree the books are good reads, but they are like \textbf{pseudo science fiction}.  & \multirow{2}{*}{\cmark}  & \multirow{2}{*}{\xmark} \\ 
                          &                    & They all agree the books are good reads, but they are \textit{too unrealistic}.              
                                                           &                     &                     \\ \hline
\multirow{6}{*}{Metaphor} & \multirow{2}{*}{$\blacktriangleright$} & The smell of smoke \textbf{carpeted} on the delinquent.                                      & \multirow{2}{*}{\xmark} & \multirow{2}{*}{\cmark}  \\ 
                          &                    & The smell of smoke \textit{took off} on the delinquent &                     &                     \\ \cline{2-5} 
                          & \multirow{2}{*}{$\blacktriangleright$} & As they strike the ground, they are \textbf{effaced}.                                        & \multirow{2}{*}{\xmark} & \multirow{2}{*}{\cmark}  \\ 
                          &                    & As they strike the ground, they are \textit{remembered}                                      &                     &                     \\ \cline{2-5} 
                          & \multirow{2}{*}{$\blacktriangleright$} & The avalanche \textbf{polvarized}  anything standing in its way.                             & \multirow{2}{*}{\xmark} & \multirow{2}{*}{\cmark}  \\ 
                          &                    & The avalanche \textit{protected}  anything standing in its way.                             
                                                                           &                     &                     \\ \hline
                          
\multirow{6}{*}{Irony} & \multirow{2}{*}{$\blacktriangleright$} & Life was never been perfect and would never be.                                      & \multirow{2}{*}{\cmark} & \multirow{2}{*}{\xmark}  \\ 
                          &                    & Life has never been perfect and would never be. &                     &                     \\ \cline{2-5} 
                          & \multirow{2}{*}{$\blacktriangleright$} & The highlight of my day figuring out how to make contact sheets \dots   such a boring life.                                        & \multirow{2}{*}{\cmark} & \multirow{2}{*}{\xmark}  \\ 
                          &                    & My entire day was occupied in making contact sheets in design such a waste.                                      &                     &                     \\ \cline{2-5} 
                                                        
                          & \multirow{2}{*}{$\blacktriangleright$} & Gotta read 70ish+ pages today \#great \#mysundayfunday \#thisshouldbefun.                                                    & \multirow{2}{*}{\xmark} & \multirow{2}{*}{\cmark}  \\ 
                          &                    & I have to read 70ish+ pages today. This is bad.                                                     &                     &                     \\ \hline
\end{tabular}
\caption{\label{tab:analysis}Examples from our Simile, Metaphor, and Irony datasets where Roberta-large fine-tuned on MNLI fails to classify the sentence pairs correctly. Gold and Pred means the true label and the predicted label respectively. $\blacktriangleright$ indicates a context and the following sentence is its corresponding hypothesis. \cmark and \xmark respectively indicate that the context entails, or does not entail the hypothesis.}
\end{table*}
\autoref{table:resultsall} reports models' accuracy on our figurative language RTE datasets.
We observe that for similes, metaphors and irony meaning, RoBERTa-large drastically outperforms the other two models. For Irony datasets, NBoW outperforms InferSent. While all models perform poorly on IIntent, InferSent's very low accuracy stands out. The low performances might be due to the templatic nature of this recast dataset which might be very different from the MNLI training data.\footnote{We leave further analysis of this issue for future work.} We now turn to an in-depth analysis of RoBERTa's performance across these datasets.

\paragraph{Ironic Meaning.}
RoBERTa-large attains over 90\% accuracy on the two datasets focused on ironic meaning. 
When analyzing these examples, a vast majority of the hypotheses in both datasets use lexical antonyms (``flattering''  $\leftrightarrow$ ``disgusting) or negation (``is great'' $\leftrightarrow$ ``is not great'') to represent the intended meaning. Thus, the presence of antonyms might be enough for RoBERTa to correctly predict that the hypothesis is not-entailed by the context.  

However, this does not hold true for hypotheses where the intended meanings were represented via more complex rephrasing. \newcite{ghosh-etal-2020-interpreting}  conducted a thorough study of the \emph{linguistic strategies} that annotators have used for the rephrasing tasks. They presented a linguistically motivated typology of the strategies (e.g., ``Lexical and phrasal antonyms'', ``Negation'', ``Weakening the intensity of sentiment'', ``Interrogative to Declarative Transformation'', ``Counterfactual Desiderative Constructions'', and ``Pragmatic Inference'') and empirically validated the strategies over the $SIGN$ and S$_{im}$-H$_{int}$ datasets.\footnote{https://github.com/debanjanghosh/interpreting\_verbal\_irony} During our analysis, 
we observe that for the vast majority of cases where RoBERTa predicts incorrectly, the examples contain Rhetorical Questions (``nice having finals on birthday?'' $\leftrightarrow$ ``do not like finals \dots''), pragmatic inferences (``Made \$174 this month \dots a yacht!"  $\leftrightarrow$ ``I don't make much money''), or desiderative constructions of \emph{[I wish] (that)} (``glad you related the news'' $\leftrightarrow$ ``[I wish] that you have told me sooner''. We also observe that  RoBERTa-large's predictions are regularly incorrect 
when the ironic messages contain certain irony markers \cite{ghosh2018marker}, such as metaphor (``shoe smell like bed of roses''  $\leftrightarrow$ ``smells bad''), alternate spelling where  the speaker  frequently  overstate  the magnitude of an ironic event (``dancing in heels is grrrrreat'' $\leftrightarrow$ ``\dots hurts your feet'') or hashtags that are composed of multi-word expressions that capture the irony (``god bless you \dots \#notinthemood).

\paragraph{Simile.} 
Likewise, for the simile dataset, we notice that RoBERTa-large often fails to reason with implicit knowledge about the physical and visual world knowledge (Table \ref{tab:analysis}). This is inline with \newcite{Weir-et-al:2020}'s finding that transformer-based contextual language models poorly capture knowledge grounded in visual perceptions. For example, RoBERTa-large incorrectly predicts that  
the context ``You wake one morning to find your entire family lying like \textbf{gray slabs of cement}'' does not entail the hypothesis ``You wake one morning to find your entire family lying \emph{unconscious}''.
Nevertheless, RoBERTa-large correctly predicts that, ``my eyes teared up \dots turning like a \textbf{ripening tomato}'' entails ``my eyes teared up \dots face \emph{turning red}''. 
 We hypothesize that here RoBERTa-large was able to identify the association between ``ripening tomato'' and ``red'' that resulted in the correct prediction. 

\paragraph{Metaphor.} We notice RoBERTa-large makes wrong predictions when it encounters \emph{unconventional} metaphors (Table \ref{tab:analysis}). Metaphors are deemed unconventional depending on ``how well-worn or how deeply entrenched a metaphor is in everyday use by ordinary people for everyday purposes" \cite{gelo2012unconventional}. For instance, for a unconventional (metaphoric, literal) pair, ``night sky \textbf{flurried} with the massive bombardment'' $\rightarrow$ ``night sky \emph{doused} with the massive bombardment'' (i.e., ``flurried'' $\leftrightarrow$ ``doused'') the model fails. On the contrary, the model correctly predicts the following conventional (metaphoric, literal) pair - ``sudden fame \textbf{kindled} her ego'' $\rightarrow$ ``\dots \emph{increased} her ego'' (i.e., ``kindled'' $\leftrightarrow$ ``increased'').

\section{Conclusion}
To understand the figurative language inference capabilities of RTE models, we introduce datasets adapted from
existing corpora focusing on similes, metaphors, and irony. By testing models trained on MNLI, we find that while the RoBERTa-large model is able to capture some aspects of figurative language, it fails when the interpretation requires word knowledge and pragmatic inferences. We hope this work will spark additional interest in the research community to incorporate and test for figurative language in their NLU systems.

\section{Ethical Considerations}
We leverage freely available open source datasets and software tools to create RTE datasets that involve similes, metaphors, and irony. We are granted the rights to further annotate and distribute the existing datasets as part of our RTE setup. This research is exempt from institutional review boards since we do not study human subjects and all social media data used is publicly available.
\bibliographystyle{acl_natbib}
\bibliography{anthology,acl2021}

\end{document}